%% file: main.tex
\documentclass[runningheads]{llncs}
\usepackage{xspace}
\usepackage{amssymb}
\newcommand{\methname}{STAER\xspace}

\usepackage{booktabs}
\usepackage{pifont}
\newcommand{\cmark}{\ding{51}} %
\newcommand{\xmark}{\ding{55}} %

\usepackage[T1]{fontenc}
\usepackage{graphicx}
\usepackage{url}
\usepackage{color}

\usepackage[table]{xcolor}
\usepackage{multirow}
\usepackage{array}
\usepackage{booktabs} %

\usepackage{colortbl}

\usepackage{algorithm}
\usepackage{algcompatible}

\usepackage{algpseudocode}
\usepackage{algorithmicx}
\algrenewcommand\textproc{} %
\algrenewcommand\algorithmiccomment[1]{\hfill{\(\triangleright\) \ #1}} %
\newcommand{\Input}{\State\textbf{Input:} }
\usepackage{setspace}
\usepackage{amsmath}

\usepackage[table]{xcolor}
\usepackage{hyperref}
\usepackage{cleveref}
\definecolor{lightblue}{RGB}{220,230,250}
\definecolor{green}{RGB}{217, 221, 146}
\newcommand{\wtxt}[1]{#1}
\newcommand{\best}[1]{\textbf{#1}}
\newcommand{\second}[1]{\underline{#1}}

\begin{document}

\title{STAER: Temporal Aligned Rehearsal for Continual Spiking Neural Network}%
\author{Matteo Gianferrari\inst{1,2}\orcidID{0009-0003-5714-4224} \and
Omayma Moussadek\inst{1,2}\orcidID{0009-0002-4714-3773} \and
Riccardo Salami\inst{1,2}\orcidID{0009-0002-0704-5810} \and
Cosimo Fiorini\inst{1,2}\orcidID{0009-0007-4083-0390} \and
Lorenzo Tartarini\inst{1,3}\orcidID{0000-0001-9976-7786} \and
Daniela Gandolfi\inst{1,3}\orcidID{0000-0003-2315-2309} \and
Simone Calderara\inst{1,2}\orcidID{0000-0001-9056-1538}}

\authorrunning{Gianferrari M. et al.}

\institute{
University of Modena and Reggio Emilia, Modena, Italy \and
AImageLab, University of Modena and Reggio Emilia, Modena, Italy \and
NILab, University of Modena and Reggio Emilia, Modena, Italy\\[2pt]
Emails: \nolinkurl{258165@studenti.unimore.it},
\nolinkurl{omayma.moussadek@unimore.it},
\nolinkurl{riccardo.salami@unimore.it},
\nolinkurl{cosimo.fiorini@unimore.it},
\nolinkurl{lorenzo.tartarini@unimore.it},
\nolinkurl{daniela.gandolfi@unimore.it},
\nolinkurl{simone.calderara@unimore.it}
}

\maketitle              %

\begin{abstract}
Spiking Neural Networks (SNNs) are inherently suited for continuous learning due to their event-driven temporal dynamics; however, their application to Class-Incremental Learning (CIL) has been hindered by catastrophic forgetting and the temporal misalignment of spike patterns. In this work, we introduce \textbf{S}piking \textbf{T}emporal \textbf{A}lignment with \textbf{E}xperience \textbf{R}eplay (\textbf{\methname}), a novel framework that explicitly preserves temporal structure to bridge the performance gap between SNNs and ANNs. Our approach integrates a differentiable Soft-DTW alignment loss to maintain spike timing fidelity and employs a temporal expansion and contraction mechanism on output logits to enforce robust representation learning. Implemented on a deep ResNet19 spiking backbone, \methname achieves state-of-the-art performance on Sequential-MNIST and Sequential-CIFAR10. Empirical results demonstrate that our method matches or outperforms strong ANN baselines (ER, DER++) while preserving biologically plausible dynamics. Ablation studies further confirm that explicit temporal alignment is critical for representational stability, positioning \methname as a scalable solution for spike-native lifelong learning. Code is available at \url{https://github.com/matteogianferrari/staer}.

\keywords{Continual Learning  \and Spiking Neural Network.}
\end{abstract}

\input{sections/01_intro}
\input{sections/02_related}

\input{sections/03_method}
\input{sections/04_experiments}

\input{sections/05_conclusions}

\bibliographystyle{splncs04}
\bibliography{bibliography_compact}
\input{supp}

\end{document}

%% file: sections/01_intro.tex
\section{Introduction}

Continual learning (CL) is a critical capability for artificial agents operating in dynamic, non-stationary environments. Rather than training on a fixed dataset, a continual learner must learn from a stream of data where new classes or tasks emerge over time. In naive approaches, updating a neural network on new data causes a drastic performance drop on previously learned classes -- a phenomenon known as \textit{catastrophic forgetting}. This challenge is particularly acute in Class-Incremental Learning (CIL), where a model must learn to distinguish all classes encountered so far without task identifiers. While methods like experience replay and knowledge distillation have enabled progress in standard Artificial Neural Networks (ANNs)~\cite{li2017learning,rebuffi2017icarl}, these models typically process information in a static, feed-forward manner. This leaves an open question: could a neural paradigm that naturally handles temporal dynamics better address the sequential nature of continual learning?

Spiking Neural Networks (SNNs) offer a compelling alternative by mimicking the event-driven processing of biological brains. Unlike ANNs, SNNs communicate via discrete spikes, encoding information in their timing and frequency, which naturally supports sequential data processing and temporal integration~\cite{roy2019towards}. However, their application to class-incremental learning remains largely unexplored and challenging. In particular, spike-based knowledge does not transfer reliably across tasks due to \textbf{temporal misalignment}: the precise spike timings encoding past classes drift as new tasks are learned, causing interference. Moreover, standard loss functions (e.g., spike counts) fail to penalize such shifts, and naive replay strategies often fail to recover the correct temporal dynamics.

In this paper, we introduce a novel SNN framework for class-incremental learning that addresses the inherent temporal challenges of continual learning in the spiking domain. We hypothesize that to prevent forgetting in SNNs, we must explicitly preserve the \textit{temporal structure} of learned representations. Inspired by biological neural computation, our approach adopts a hybrid coding strategy in which information is jointly encoded by firing rates and spike timing. This dual coding principle is central to biological information processing and suggests that maintaining both rate-based and temporal information is crucial for robust and adaptive behavior in spiking neural networks. More specifically, the rate-based coding hypothesis states that information is encoded in the average firing rate of neurons over a given time window, and has been extensively supported by experimental findings across sensory and motor cortices~\cite{shadlen1994noise}. This paradigm underlies many classical models of neural computation and learning, including Hebbian plasticity mechanisms, where synaptic strength is modulated by correlated firing activity~\cite{hebb1949organization}. In parallel, it is well established that the temporal structure of spike trains, including precise spike timing and relative spike latencies, can carry significant information beyond firing rates alone. Temporal coding schemes have been observed in various neural systems, such as the auditory pathway, hippocampus, and visual cortex and cerebellum, where millisecond-scale spike timing conveys stimulus-specific or contextual information~\cite{thorpe2001spike}. This view is strongly supported by the discovery of spike-timing-dependent plasticity (STDP), a biologically plausible learning rule in which synaptic modifications depend on the relative timing of pre- and post-synaptic spikes\cite{caporale2008spike}. To this end, our method integrates two biologically inspired mechanisms. First, we incorporate a differentiable temporal alignment loss based on \textbf{Soft Dynamic Time Warping (soft-DTW)}~\cite{cuturi2017soft,blondel2021differentiable}.
By applying soft-DTW between current and past output sequences during replay, we ensure that the network maintains the temporal fidelity of its responses across tasks. Second, by leveraging experience-replay, we \textbf{expand and contract past responses} along the temporal axis and align them with the responses of the current network to enforce recoverability of past behavior under different temporal samplings. %

We implement our approach on a deep convolutional spiking architecture, specifically a ResNet19 SNN~\cite{zheng2021going}, %
to ensure scalability to complex image benchmarks. Unlike prior SNN-CL works that often rely on shallow networks or static datasets, our extensive evaluation on class-incremental benchmarks (e.g., Sequential-CIFAR10, Sequential MNIST) demonstrates that deep SNNs can achieve accuracy competitive with state-of-the-art ANN baselines.

Our main contributions are summarized as follows:
\begin{itemize}
    \item \textbf{Novel Temporal Alignment Framework:} We propose the first SNN-CL method that utilizes differentiable soft-DTW to explicitly align spike timings between tasks, mitigating the temporal drift that causes catastrophic forgetting. %
    \item \textbf{State-of-the-Art Performance:} We show that our alignment-enhanced ResNet19 SNN achieves performance on par with strong ANN baselines in class-incremental settings, bridging the accuracy gap between neuromorphic and traditional continual learning.
\end{itemize}

%% file: sections/02_related.tex
\section{Related works}
ResNet is one of the most popular architectures to tackle the problem of degradation in deeper neural networks~\cite{lee2024impactmodelsizecatastrophic}. Such networks explicitly reformulate the layers as learning residual functions with reference to the layer inputs, instead of learning unreferenced functions~\cite{he2016deep}. ResNet has been widely employed as a backbone network in continual learning benchmarks due to its strong empirical performance, stable optimization properties and widespread adoption in image classification tasks~\cite{buzzega2020dark,XU2026112213}. 
As most ANN models, ResNet suffers from catastrophic forgetting, whereby learning a new task leads to a degradation of performance on previously acquired tasks~\cite{mccloskey1989catastrophic,lee2024impactmodelsizecatastrophic,rebuffi2017icarl}. This limitation motivates the investigation of alternative neural paradigms inspired by biological computation. In this context, SNN encode neuronal activity through sparse, binary communication, mimicking the physiological synaptic dynamics of the human brain~\cite{roy2019towards}. Such properties offer mechanisms that may naturally mitigate catastrophic forgetting by enabling new knowledge to be built upon previously learned representations. In biological systems, long-term memory formation emerges from the interaction of multiple mechanisms, including synaptic plasticity, memory replay, neuromodulatory signals, and recurrent connectivity~\cite{kopsick2024formation}. Recent SNN models have been explicitly designed to reproduce these effects. For instance, in SNNs synaptic weight updates occur only when a presynaptic neuron’s membrane potential reaches a threshold and emits a spike, in contrast to ANNs where weights are updated even for neurons with very small activation values~\cite{PARISI201954,ONEILL2010220}. This event-driven learning mechanism has been suggested to act as a form of task-dependent gating, whereby synapses critical for previously learned tasks are preserved when the corresponding neurons remain inactive during the learning of new tasks, alleviating catastrophic forgetting. Building on this idea, a neuromodulatory network that dynamically adapts neuronal firing thresholds based on the spiking activity of the preceding layer has been shown to further enhance memory retention~\cite{hammouamri2022mitigating}. Additionally, memory consolidation is supported by memory replay, which refers to the reactivation of previously experienced episodes~\cite{ROSCOW2021808}. Notably, it has been hypothesized that temporally compressed or reversed replay of events plays a central role in this process~\cite{nicola2017force,euston2007fastforward}. Experimental studies have reported the presence of both temporal compression and dilation in the striatum~\cite{mello2015scalable,motanis2015time}. While temporal compression during memory replay has been successfully modeled and reproduced in SNNs~\cite{nicola2017force}, the potential role of neural time dilation remains largely unexplored. Multiple studies further highlight that during both learning and sleep, firing activity patterns are temporally compressed. For instance, in rodents, hippocampal place cells—representing specific positions along the animal’s path—reactivate on a faster timescale than during actual navigation, reflecting a compressed replay of past trajectories~\cite{diba2007forward}. 

Nevertheless, due to their highly nonlinear and spiking nature, training SNN remains challenging. Significant research efforts have focused on leveraging local, biologically plausible learning signals—such as Hebbian learning rules~\cite{morawiecki2022hebbCL,hebb1949organization} and Spike-Timing-Dependent Plasticity (STDP)~\cite{Allred2020STDP,Markram2012STDP} —to enable unsupervised training based on the causal relationships between the firing activities of connected neurons.
In addition, eligibility trace mechanisms have been explored to propagate error signals and approximate backpropagation by attributing the network loss to individual synaptic contributions~\cite{bellec2020learningdilemma}. Despite these limitations, directly trained deep SNNs based on the ResNet architecture~\cite{zheng2021going} using surrogate gradient descent~\cite{zimmer2019snnpytorch} for backpropagation have demonstrated competitive—and in several benchmarks, superior—performance on datasets such as CIFAR-10 and ImageNet, outperforming approaches based on ANN-to-SNN conversion, spike-based backpropagation, and hybrid  training approach~\cite{zheng2021going}.

To enable deep SNNs to model memory replay with time compression and dilation, a mathematical tool for comparing spike trains with variable temporal structure is needed. Soft-DTW has been introduced as a differentiable loss function to compare time series of variable size, robust to shifts or dilatations across the time dimension~\cite{cuturi2017soft}. Such loss term is affine to the recall of temporal firing patterns of neurons, when a previously learned stimuli is represented in its entirety or fractionally.

%% file: sections/03_method.tex
\input{figures/mainfig}
\input{figures/backbone}
\input{algorithms/main_algo}

\section{Methodology}

We formalize continual learning for spiking neural networks as follows. Let  $\mathcal{T} = \{1, \dots, N\}$  denote the set of tasks, where each task  $n \in \mathcal{T}$  provides samples $(x_n, y_n)$ drawn from distribution $\mathcal{D}_n$. The model is represented by a function $f$. For an input $x$, the network at current task $n$ produces temporal logits $h_n^T(x) \in \mathbb{R}^{T \times C}$ , where $T$ is the number of time steps and $C$ the number of classes. Probability distribution over classes is defined as  $ f(x) = \text{softmax}\big( \bar{h}_n(x) \big) $, where $\bar{h}_n(x)$ represents the average over time of logits values. Our goal is to learn $f$ sequentially across tasks avoiding catastrophic forgetting, while preserving both semantic and temporal structure in spike-based representations.

\subsection{Preliminaries}

\paragraph{\textbf{Spiking Neuron Model}}
\label{sec:surrogate_gradient}
Following recent directly-trained SNN works~\cite{yu2025temporal,zheng2021going}, we model each layer with discrete-time Leaky Integrate-and-Fire (LIF) neurons. For a given layer $\ell$, at time step $t$, the synaptic current $\mathbf{I}^{\ell}[t]$ and membrane potential $\mathbf{u}^{\ell}[t]$ evolve as
\begin{equation}
\mathbf{I}^{\ell}[t] = \mathbf{W}^{\ell}\,\mathbf{s}^{\ell-1}[t], \qquad
\mathbf{u}^{\ell}[t] = \gamma\,\mathbf{u}^{\ell}[t-1] + \mathbf{I}^{\ell}[t] - \mathbf{s}^{\ell}[t-1]\,V_{\mathrm{th}},
\end{equation}
where $\mathbf{W}^\ell$ are the learned weights, $\mathbf{s}^{\ell}[t]\in\{0,1\}^{m_\ell}$ is the layer-$\ell$ spike vector (one component per neuron), with $m_\ell$ neurons, $\gamma\in(0,1)$ is the leak coefficient, and $V_{\mathrm{th}}$ is the firing threshold. Spikes are detected by thresholding the membrane potential through a Heaviside firing function
\begin{equation}
\mathbf{s}^{\ell}[t] = H\!\left(\mathbf{u}^{\ell}[t]-V_{\mathrm{th}}\right).
\end{equation}
The subtractive term $\mathbf{s}^{\ell}[t-1]V_{\mathrm{th}}$ applies a soft reset after the onset of a spike.

Since $H(\cdot)$ is non-differentiable, we use surrogate gradients (as implemented in \textit{snnTorch}, adapted from~\cite{fang2021incorporating}), keeping the hard threshold in the forward pass and approximating the backward derivative as \begin{equation}
\frac{\partial \mathbf{s}}{\partial \mathbf{u}} \approx g(\mathbf{u}-V_{\mathrm{th}}), \;\;\;\text{where} \;\;\; g(z)=\frac{\alpha}{2}\,\frac{1}{1+\left(\frac{\pi\alpha}{2}z\right)^2}
\end{equation}
is the arctangent surrogate with slope parameter $\alpha$.

\paragraph{\textbf{Continual Learning and Experience Replay}}

A widely adopted method to address catastrophic forgetting is Experience Replay (ER)~\cite{ratcliff1990connectionist,robins1995catastrophic}, known for both its simplicity and efficacy. It maintains a memory buffer $\mathcal{B}$ that stores a subset of samples from previous tasks, along with their labels. During training on task $n$, mini-batches drawn from $\mathcal{B}$ are interleaved with current data, allowing the model to rehearse past knowledge while learning new classes.
Dark Experience Replay (DER) and its extension  DER++~\cite{buzzega2020dark} improve stability by applying distillation on buffer data and avoiding overfitting on stored logits. Specifically, DER extends ER by also storing the model’s output logits for replayed samples. This enables the network to preserve its previous predictions, reducing representation drift across tasks. DER++ further improves this approach by introducing an additional loss term for buffer data to lower the impact of  highly biased logits learned in previous tasks. These two method represent simple yet strong ANN baselines in class-incremental settings, which we compare with our solution.
In our framework, the buffer $\mathcal{B}$ is adapted to the spiking domain: it retains input samples and temporal spike-based outputs that are further augmented through a time contraction and expansion mechanism.

\paragraph{\textbf{Temporal Alignment with Soft-DTW}}

Dynamic Time Warping (DTW)~\cite{berndt1994DTW} is a classical technique for measuring similarity between time series by computing an optimal alignment that accounts for local shifts along the temporal axis. However, its non-differentiable nature limits its integration into gradient-based models. Soft-DTW, introduced by~\cite{cuturi2017soft}, provides a differentiable relaxation of DTW by replacing the hard minimum operator with a soft-min, enabling smooth optimization while preserving robustness to temporal distortions. This property has made soft-DTW widely adopted in machine learning for tasks such as sequence classification and representation learning, where maintaining temporal structure is critical~\cite{blondel2021differentiable}.

\subsection{Spiking Temporal Alignment with Experience Replay}
\label{sec:static_encoding}
Unlike standard SNNs, \methname outputs real-valued logits rather than spike counts, following the strategy introduced in~\cite{yu2025temporal}. This design enables compatibility with conventional loss functions and facilitates temporal alignment across tasks. We adopt a ResNet19 SNN backbone, as seen in~\Cref{fig:backbone}, trained with surrogate gradients, which can operate under different temporal resolutions. Specifically, once trained, the network can process inputs over variable time windows $T$, allowing flexible temporal sampling without retraining.

For each sample $x$ stored in the replay buffer, we maintain its related label $y$, and three sequences of logits corresponding to time windows $T$, $T/2$, and $2T$. In our setting, the input is augmented through a \textit{static encoding} that replicates the same image $T$ times to form the temporal axis. Consequently, temporal contraction and expansion are obtained by decreasing or increasing the replication length $T$ in the static encoding, yielding sequences of length $T/2$ and $2T$, respectively. This temporal expansion and contraction mechanism is inspired by biological memory processes~\cite{mello2015scalable,motanis2015time}, where recall traces can undergo compression or dilation over time.

As described in \Cref{fig:model}, the learning objective combines two components. First, we employ the standard cross-entropy loss $\mathcal{L}_{CE}$ on a batch obtained by concatenating samples from the current task with samples retrieved from the replay buffer. The loss is computed on logits averaged along the temporal dimension.
Given the previously defined samples and labels from the current task $n$, and let $(x_{\mathcal{B}}, y_{\mathcal{B}})$ be those retrieved from the buffer $\mathcal{B}$, we then form the concatenated batch
\begin{equation}
x = \left[ x_n; x_{\mathcal{B}} \right], \qquad
y = \left[ y_n; y_{\mathcal{B}} \right],
\end{equation}
and compute the cross-entropy loss over the combined set as
\begin{equation}
\mathcal{L}_{CE} = \mathrm{CE}\!\left(\bar{h}_n(x), y \right).
\end{equation}

For replayed samples, we introduce a Temporal Alignment loss $\mathcal{L}_{TA}$ based on Soft-DTW (SDTW), which enforces consistency between the temporal structure of current and past outputs. Let $h_n^T(x_{\mathcal{B}})\in\mathbb{R}^{T\times C}$ denote the current network output (logits) at task $n$ over $T$ time steps. For the same replay sample $x_{\mathcal{B}}$, the buffer stores three logit sequences computed by a past model snapshot at different temporal resolutions:
$h_{\mathcal{B}}^{T}\in\mathbb{R}^{T\times C}$,
$h_{\mathcal{B}}^{T/2}\in\mathbb{R}^{(T/2)\times C}$,
and $h_{\mathcal{B}}^{2T}\in\mathbb{R}^{(2T)\times C}$.
Since Soft-DTW naturally aligns sequences of unequal length, we can compare the current $T$-step logits to each stored sequence, obtaining three SDTW terms:
\begin{align}
\mathcal{L}^{T}_{\mathrm{SDTW}}(x_{\mathcal{B}})   &= \mathcal{L}_{\mathrm{SDTW}}\!\Big(h_{n}^{T}(x_{\mathcal{B}}),\,h_{\mathcal{B}}^{T}\Big),\\
\mathcal{L}^{T/2}_{\mathrm{SDTW}}(x_{\mathcal{B}}) &= \mathcal{L}_{\mathrm{SDTW}}\!\Big(h_{n}^{T}(x_{\mathcal{B}}),\,h_{\mathcal{B}}^{T/2}\Big),\\
\mathcal{L}^{2T}_{\mathrm{SDTW}}(x_{\mathcal{B}})  &= \mathcal{L}_{\mathrm{SDTW}}\!\Big(h_{n}^{T}(x_{\mathcal{B}}),\,h_{\mathcal{B}}^{2T}\Big).
\end{align}
The temporal-alignment loss is the normalized weighted combination:
\begin{equation}
\mathcal{L}_{TA} =
\frac{\mathcal{L}^{T}_{\mathrm{SDTW}} + \alpha_1\,\mathcal{L}^{T/2}_{\mathrm{SDTW}} + \alpha_2\,\mathcal{L}^{2T}_{\mathrm{SDTW}}}{1+\alpha_1+\alpha_2},
\end{equation}
where $\alpha_1$ and $\alpha_2$ are hyperparameters weighting the contribution of contraction and expansion terms.
The overall training objective is expressed as:
\begin{equation}
    \mathcal{L} = \mathcal{L}_{CE} + \beta \cdot \mathcal{L}_{TA}
\end{equation}
where $\beta$ balances current-task learning and temporal alignment.
The method is described in detail in \Cref{algo:model}.

%% file: figures/mainfig.tex
\begin{figure}[t]
    \centering
    \includegraphics[width=1\textwidth]{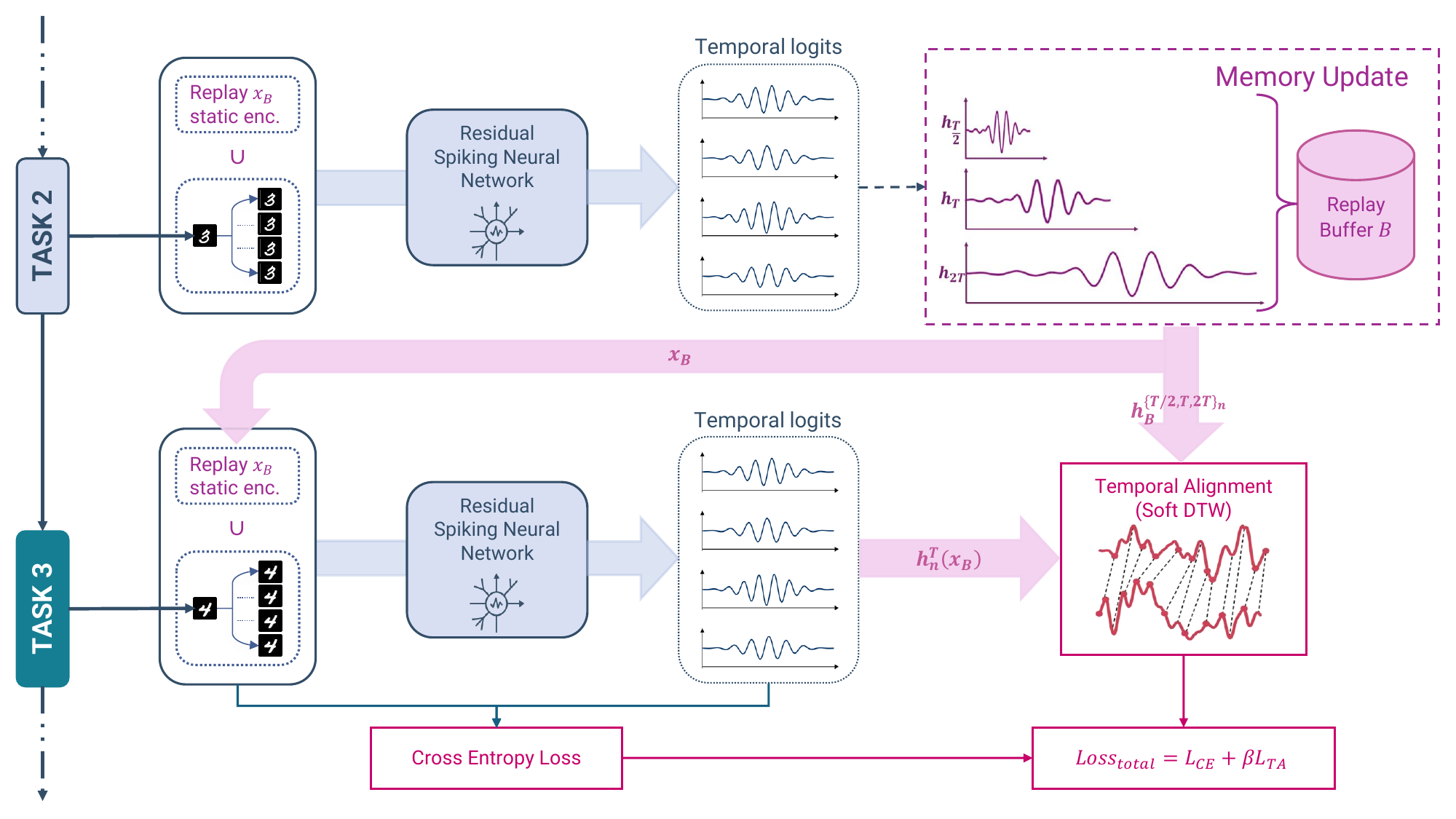}
    \caption{For each input, logits are stored at three temporal resolutions $(T, T/2, 2T)$ in the replay buffer to mimic biological memory variability. The training objective combines cross-entropy on current task samples with a Temporal Alignment (TA) loss based on Soft-DTW, computed between current and past logits at multiple temporal scales.}
    \label{fig:model}
\end{figure}

%% file: figures/backbone.tex
\begin{figure}[t]
    \centering
    \includegraphics[width=1\textwidth]{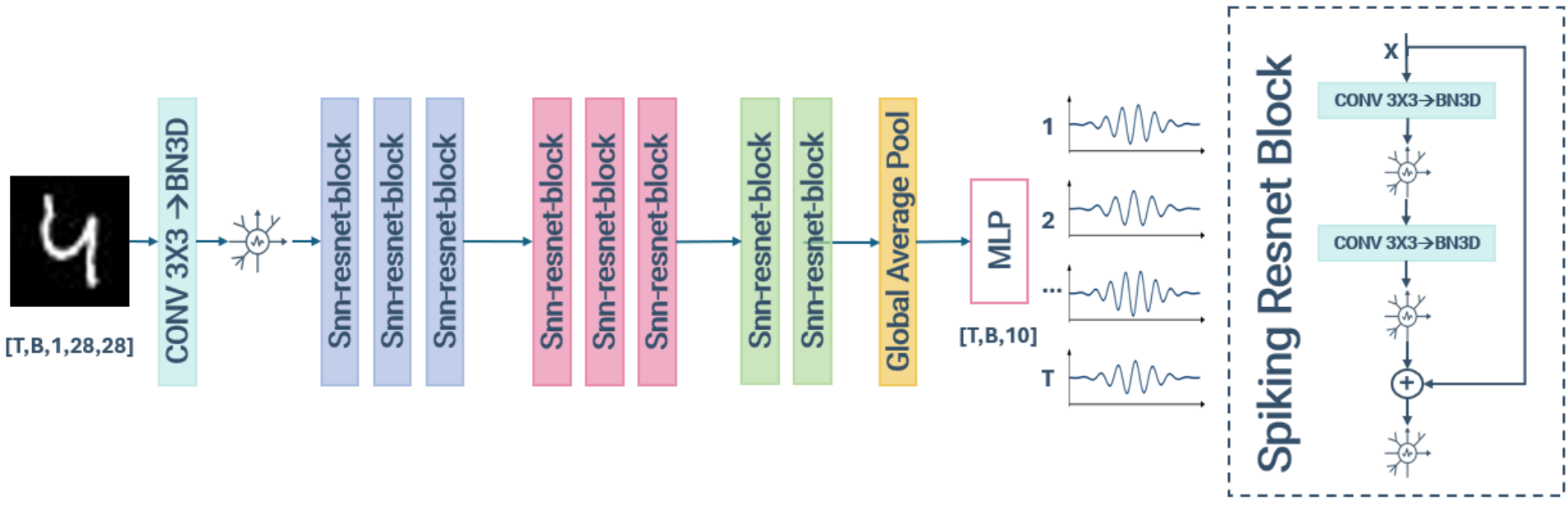}
    \caption{\textit{Spiking ResNet19 backbone.} \textit{Left:} A static input image is replicated over $T$ time steps to form a $[T, B, C, H, W]$ tensor, then processed by an initial convolutional stem followed by a sequence of spiking residual blocks. The network ends with global average pooling and an MLP head, producing real-valued logits at every time step ($[T, B, 10]$ for our datasets). \textit{Right:} Detail of a spiking ResNet block: the main branch stacks convolution--normalization layers interleaved with LIF neurons, while a residual shortcut is added to the main branch output before the next spiking activation.}

    \label{fig:backbone}
\end{figure}

%% file: algorithms/main_algo.tex
\algnewcommand\algorithmicinput{\textbf{Input:}}
\algnewcommand\Input{\item[\algorithmicinput]}

\begin{algorithm}[t]
\caption{\methname\ algorithm}
\label{algo:model}
{\setstretch{1.3}
\begin{algorithmic}[1]
\Input $(x_n,y_n)\in\mathcal{D}_n$ new task-$n$ data; $h_n^{T}(x)$ logits of the model trained up to task $n$ with time window $T$; 
$(x_{\mathcal{B}},y_{\mathcal{B}}, h_{\mathcal{B}}^{T/2}, h_{\mathcal{B}}^{T}, h_{\mathcal{B}}^{2T}) \in \mathcal{B}$ buffer contents.

\State $\mathcal{B} \gets \emptyset$

\For{$n \in \{1,\ldots,N\}$}
    \State \underline{Training}
    \State $\mathcal{L} \gets \mathcal{L}_{CE}\!\Big(\bar{h}_n\big([x_n,x_{\mathcal{B}}]\big), [y_n,y_{\mathcal{B}}]\Big)$ \Comment{CE on new+buffer}
    \State $\mathcal{L} \gets \mathcal{L} + \beta\,\mathcal{L}_{TA}\!\Big(h_n^{T}(x_{\mathcal{B}}), h_{\mathcal{B}}^{T/2}, h_{\mathcal{B}}^{T}, h_{\mathcal{B}}^{2T}\Big)$ \Comment{Temporal alignment}

    \State \underline{Update buffer}
    \State $x_{\mathcal{B}} \gets x_n,\quad y_{\mathcal{B}} \gets y_n$ \Comment{Update examples}
    \State $%
           \mathcal{B} \gets (h_n^{T/2}(x_{\mathcal{B}}),\; h_n^{T}(x_{\mathcal{B}}),\; h_n^{2T}(x_{\mathcal{B}})$) \Comment{Update multi-time logits}
\EndFor
\end{algorithmic}
}
\end{algorithm}

%% file: sections/04_experiments.tex
\input{figures/ablation}
\section{Experiments}
\label{sec:exp}
\paragraph{\textbf{Datasets.}}

We evaluate \methname\ on two standard continual learning benchmarks, Sequential-MNIST and Sequential-CIFAR10~\cite{zenke2017continual}, each composed of 10 classes split into 5 sequential tasks with 2 new classes per task.
Since \methname\ operates on temporal inputs while ANN baselines use static images, we apply a \textit{static encoding} strategy (Sec.~\ref{sec:static_encoding}) by replicating each image over $T$ time steps, yielding a tensor of shape $T \times B \times C \times H \times W$, where $T$ is the number of time steps, $B$ the batch size, $C$ the number of channels, and $H \times W$ the spatial resolution.
Sequential-MNIST contains $28 \times 28$ grayscale digit images, while Sequential-CIFAR10 consists of $32 \times 32$ RGB images with standard data augmentation (random cropping, horizontal flipping, and normalization).

\paragraph{\textbf{Implementation details}}
To ensure a fair comparison, both ANN and SNN use the same ResNet19 backbone, differing only in neuron model and training dynamics, denoted as \textit{ResNet19} (ANN) and \textit{Spiking ResNet19} (SNN), following prior SNN literature~\cite{yu2025temporal,zheng2021going,deng2022temporal,yao2022glif}.

The architecture consists of a convolutional stem with 128 channels followed by three residual stages with 3, 3, and 2 Basic Blocks, with channel widths $\{128,256,512\}$ and downsampling at the start of the last two stages. Classification uses global average pooling and a linear layer. The ANN employs ReLU with 2D BatchNorm, while the SNN uses LIF neurons with $\gamma=0.5$ and $V_{th}=1.0$, trained via ArcTan surrogate gradients (Sec.~\ref{sec:surrogate_gradient}).

For $\mathcal{L}_{TA}$, the contraction and expansion terms $\mathcal{L}_{\mathrm{SDTW}}^{T/2}$ and $\mathcal{L}_{\mathrm{SDTW}}^{2T}$ are computed on half of each mini-batch. We use replay buffers $B\in\{200,500,5120\}$, train 50 epochs per task on Sequential-CIFAR10 (batch size 32) and 1 epoch per task on Sequential-MNIST (batch size 10), with $T\in\{2,4\}$. We set $\alpha_1=\alpha_2=0.5$, $\beta=10^{-4}$, and use Adam with learning rate $3\times10^{-3}$ and a cosine scheduler.

\paragraph{\textbf{Evaluated approaches}}
We compare our method with replay-based baselines and two reference bounds, reporting results in both ANN and SNN settings.
As a lower bound, we use \textit{SGD} in a sequential fine-tuning without any continual mechanism, while \textit{Joint} training on the union of all tasks serves as an upper bound.
We additionally evaluate \textit{ER}~\cite{riemer2018learning}, which mitigates forgetting by replaying a fixed-size buffer of past samples during training. We also consider \textit{DER} and \textit{DER++}~\cite{buzzega2020dark}, which extend ER by storing past logits and distilling them during replay, with DER++ further adding a supervised loss on buffered labels to improve stability, especially for small buffers.
Since these baselines originally proposed and evaluated in the ANN literature, to ensure a fair comparison across paradigms, we align the network capacity by changing the ANN baselines to use the same backbone as our spiking model: specifically, we replace their default ResNet18 backbone with a ResNet19 
counterpart matching our SNN  backbone.

As in many continual learning works, we evaluate our method under standard protocols that differ in how tasks are defined and which information is available at test time.
Following the three main continual learning scenarios~\cite{van2019three}, we perform  Task-Incremental Learning (TIL) and  Class-Incremental Learning (CIL) experiments.
TIL is typically easier because the task identity is provided at inference time.
Our main focus is CIL, where the model is trained sequentially on disjoint class subsets and must, at test time, recognize samples among all classes observed so far without access to task identity.

\paragraph{\textbf{Evaluation metrics}}
~\Cref{tab:main_table} reports results in terms of \textit{Final Average Accuracy} (FAA) which measures the average classification accuracy across all tasks after the model has completed the entire incremental training process. Let $R_{N,n}$ denote the accuracy on task $n$ evaluated after training on the final task $N$. The FAA is then defined as:
\begin{equation}
\mathrm{FAA} = \frac{1}{N} \sum_{n=1}^{N} R_{N,n}.
\end{equation}
FAA does not explicitly quantify how much performance on earlier tasks degrades during the incremental process, we therefore also report in~\Cref{tab:forgetting} \textit{Forgetting} (FRG), computed from the accuracy matrix $\{R_{k,n}\}$, where $R_{k,n}$ is the accuracy on task $n$ after training up to task $k$. For each task $n<N$, forgetting is measured as the drop between the best accuracy ever achieved on that task and its final accuracy after learning all tasks. We aggregate it as:
\begin{equation}
\mathrm{FRG}=\frac{1}{N-1}\sum_{n=1}^{N-1}\Big(\max_{k\in\{n,\dots,N\}} R_{k,n}-R_{N,n}\Big),
\end{equation}
where lower values indicate better retention.

\paragraph{\textbf{Results}}
\input{tables/main_table}
Our method consistently improves upon spiking replay baselines across both datasets. In particular, with \(T=4\) we achieve the best SNN performance for all buffer sizes on Sequential-MNIST (up to \(98.48\%\)) and Sequential-CIFAR10 (up to \(83.53\%\)), outperforming snn-ER and providing clear gains over snn-DER / snn-DER++. With \(T=2\), \methname remains competitive on Sequential-MNIST and improves over snn-ER, while on the more challenging Sequential-CIFAR10 the advantages become less consistent for large buffers.
The temporal alignment becomes more effective as the number of time steps increases, with the largest improvements observed on Sequential-CIFAR10 (e.g \(+6.56\%\) at \(B{=}500\)). This shows the effectiveness of our method, while for snn-ER, and snn-DER / snn-DER++, the increase in time steps does not impact the performance as much.

\paragraph{\textbf{Ablation studies}}
We ablate the Temporal Alignment (TA) objective $\mathcal{L}_{TA}$ of \methname by sweeping the alignment weight $\beta\in \{10^{-4},10^{-3},5\times10^{-3}\}$ and the Soft-DTW compression/dilation weights $\alpha_1,\alpha_2\in \{0,0.1,0.25,0.5,0.75,1.0\}$ and $T=2$, and report the final CIL accuracy on Sequential-MNIST, averaged over 5 random seeds.~\Cref{fig:ablation} shows that the best configuration is obtained with a moderate and balanced alignment, reaching $93.71\%$ for $\beta=10^{-4}$ and $\alpha_1=\alpha_2=0.5$. For $\beta=10^{-4}$, the performance landscape is comparatively smooth, but it becomes noticeably worse when dilation is over-emphasized (large $\alpha_2$) relative to compression. In contrast, $\beta=10^{-3}$ yields a much more brittle regime: while the best setting attains $93.32\%$ at $(\alpha_1,\alpha_2)=(0.1,0)$, several configurations collapse severely (down to $24.69\%$), validating that over-emphasized dilation over compression results in degradation. Increasing the weight to $\beta=5\times10^{-3}$ recovers strong peak performance (up to $92.29\%$ at $\alpha_1=\alpha_2=1.0$), although unbalanced settings can still degrade substantially. Finally, setting $\alpha_1=\alpha_2=0$ (i.e., removing the contraction/expansion SDTW terms and keeping only same-resolution alignment) reduces the FAA to $89.95\%$ for $\beta=10^{-4}$, supporting the utility of the additional temporal resolutions.~\Cref{tab:ablation_table} shows the contribution of each component in loss $\mathcal{L}_{TA}$. Based on this study, we fix $\beta=10^{-4}$ and $\alpha_1=\alpha_2=0.5$ in all subsequent experiments.

\input{tables/ablation_table}

\paragraph{\textbf{Forgetting results}}
As expected, increasing the replay buffer consistently reduces forgetting for every method, with the effect being particularly pronounced in the Sequential-CIFAR10 setting. Increasing the number of time steps from $T=2$ to $T=4$ systematically improves SNN stability for all methods. Overall, \methname yields consistently the lowest forgetting in most configurations, even surpassing the ANN models in the most constrained setting, showing that SNN are naturally better at handling forgetting.

\input{tables/forgetting.tex}

%% file: figures/ablation.tex
\begin{figure}[t]
    \centering
    \includegraphics[width=1\textwidth]{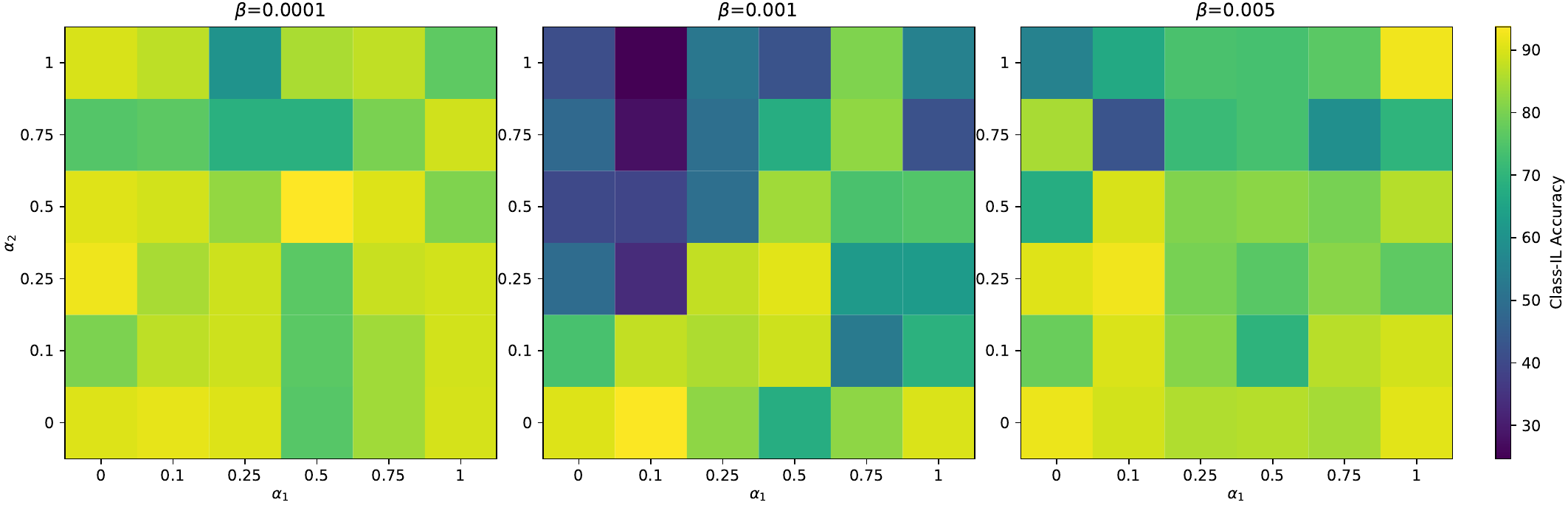}
    \caption{Hyperparameter sensitivity of the Temporal Alignment (TA) objective on Sequential-MNIST. Each heatmap (fixed $\beta$) reports the final CIL accuracy (lighter color is better) as a function of the SDTW compression/dilation weights $(\alpha_1,\alpha_2)$ and the TA strength $\beta$. The x-axis shows $\alpha_1$ (compression), the y-axis shows $\alpha_2$ (dilation).}
    \label{fig:ablation}
\end{figure}

%% file: tables/main_table.tex
\begin{table*}[tb]
\caption{Results on Sequential-MNIST and Sequential-CIFAR10 under CIL.
Bold and underlined values denote the best and second-best results within each SNN block (T=2 and T=4), respectively.}
\label{tab:main_table}

\centering
\small
\setlength{\tabcolsep}{6pt}
\renewcommand{\arraystretch}{1.2}

\begin{tabular}{lcccccc}
\toprule
\textbf{Method} &
\multicolumn{3}{c}{\textbf{S-MNIST}} &
\multicolumn{3}{c}{\textbf{S-CIFAR-10}} \\
\textit{Buffer size} 
 &
\textit{200} & \textit{500} & \textit{5120} &
\textit{200} & \textit{500} & \textit{5120} \\
\midrule

\rowcolor{green}
\multicolumn{7}{c}{\textbf{ANN without T}} \\

JOINT  & & 99.32& &  & 98.25 &  \\
SGD    & & 19.89&  & & 19.63 &  \\
\midrule
ER     & 94.46 & 95.39 & 98.17 & 46.68 & 61.64 & 85.91 \\
DER    & 96.12 & 97.91 & 98.83 & 51.71 & 64.51 & 86.27 \\
DER++  & 96.37 & 98.15 & 99.06 & 56.57 & 65.94 & 87.08 \\

\midrule
\rowcolor{green}
\multicolumn{7}{c}{\textbf{SNN with T=2}} \\

snn-JOINT & & 98.72 & & & 89.75 &  \\
snn-SGD   & & 19.64 & & & 19.44 &  \\
\midrule
snn-ER     & 87.90 & 94.06 & 95.20 & 37.80 & 52.68 & 72.89 \\
snn-DER    & 92.54 & 94.88 & 96.88 & \second{48.56} & 58.62 & 75.92 \\
snn-DER++  & \second{93.54} & \second{96.01} & \second{97.33} & \best{50.40} & \second{59.08} & \second{78.19} \\

\midrule
\rowcolor{lightblue}
\textbf{\methname} & \best{\wtxt{93.71}} & \best{\wtxt{96.70}} & \best{\wtxt{97.53}} & \wtxt{47.88} & \best{\wtxt{59.12}} & \best{\wtxt{78.30}} \\

\midrule
\rowcolor{green}
\multicolumn{7}{c}{\textbf{SNN with T=4}} \\

snn-JOINT & & 98.92 & & & 89.80 &  \\
snn-SGD   & & 19.68 & & & 19.47 &  \\
\midrule
snn-ER     & 91.20 & 94.00 & 96.55 & 38.99 & 50.82 & 74.44 \\
snn-DER    & \second{94.07} & 95.79 & 97.78 & 50.16 & 57.24 & 75.80 \\
snn-DER++  & 93.92 & \second{96.00} & \second{98.04} & \second{51.11} & \second{60.81} & \second{78.25} \\

\rowcolor{lightblue}
\textbf{\methname} & \best{\wtxt{94.18}} & \best{\wtxt{96.92}} & \best{\wtxt{98.48}} & \best{\wtxt{51.19}} & \best{\wtxt{65.68}} & \best{\wtxt{83.53}} \\

\bottomrule
\end{tabular}
\end{table*}

%% file: tables/ablation_table.tex
\begin{table*}[t]
\caption{CIL accuracy on Sequential-MNIST enabling/disabling individual loss components to quantify their contribution.}
\label{tab:ablation_table}
\centering
\small
\setlength{\tabcolsep}{6pt}
\renewcommand{\arraystretch}{1.2}
\begin{tabular}{lcccc}
\toprule
\textbf{Setting} &
$\boldsymbol{\beta}$  &
$\boldsymbol{\alpha_1}$ &
$\boldsymbol{\alpha_2}$ &
\textbf{Class-IL} \\
\midrule
snn-ER & \xmark & \xmark & \xmark & 87.90 \\
snn-ER + SDTW & \cmark & \xmark & \xmark & 90.15 \\
\methname w/o dilation & \cmark & \cmark & \xmark & 91.02 \\
\methname w/o compression & \cmark & \xmark & \cmark & 91.69 \\
\textbf{\methname} & \cmark & \cmark & \cmark & \best{\wtxt{93.71}} \\
\bottomrule
\end{tabular}
\end{table*}

%% file: tables/forgetting.tex
\begin{table*}[tb]
\caption{Forgetting results on Sequential-MNIST and Sequential-CIFAR10 under CIL. Bold and underlined values denote the best and second-best results within each SNN block (T=2 and T=4), respectively.}
\label{tab:forgetting}

\centering
\small
\setlength{\tabcolsep}{6pt}
\renewcommand{\arraystretch}{1.2}

\begin{tabular}{lcccccc}
\toprule
\multicolumn{7}{c}{\textbf{FORGETTING}} \\
\midrule

\textbf{Method} &
\multicolumn{3}{c}{\textbf{S-MNIST}} &
\multicolumn{3}{c}{\textbf{S-CIFAR-10}} \\
\textit{Buffer size} 
 &
\textit{200} & \textit{500} & \textit{5120} &
\textit{200} & \textit{500} & \textit{5120} \\
\midrule

\rowcolor{green}
\multicolumn{7}{c}{\textbf{ANN without T}} \\

ER     & 6.10 & 5.33 & 1.46 & 63.11 & 44.66 & 17.15 \\
DER    & 4.46 & 1.57 & 0.66 & 55.85 & 44.23 & 13.47 \\
DER++  & 4.04 & 1.94 & 0.04 & 47.81 & 36.41 & 11.60 \\

\midrule
\rowcolor{green}
\multicolumn{7}{c}{\textbf{SNN with T=2}} \\

snn-ER    & 14.40 & 6.03 & 4.88 & 71.72 & 55.52 & 19.83 \\
snn-DER   & 6.97 & 2.27  & 1.77 & 47.80 & 35.38 & 18.78 \\
snn-DER++ & \second{5.07} & \second{1.92} & \second{1.05} & \second{42.41}  & \best{31.86} & \second{15.61} \\

\midrule
\rowcolor{lightblue}
\textbf{\methname} & \best{4.05} & \best{1.70} & \best{0.67} & \best{41.91} & \second{33.68} & \best{13.85} \\

\midrule
\rowcolor{green}
\multicolumn{7}{c}{\textbf{SNN with T=4}} \\

snn-ER    & 10.16 & 4.23 & 1.82 & 70.31 & 53.2 & 19.37 \\
snn-DER   &  4.82 & 2.08 & 0.92 & 38.55 &  31.83 & \second{13.39} \\
snn-DER++ &  \second{4.27} & \second{1.44} & \second{0.80} & \best{32.66} & \second{31.42} & 13.46 \\

\rowcolor{lightblue}
\textbf{\methname} & \best{2.14} & \best{0.94} & \best{0.44} & \second{35.47} & \best{29.20} & \best{11.45} \\

\bottomrule
\end{tabular}
\end{table*}

%% file: sections/05_conclusions.tex
\section{Conclusions}
\vspace{-0.2em}
Spiking Neural Networks (SNNs) offer a compelling foundation for lifelong learning due to their intrinsic temporal processing; however, their potential in class-incremental scenarios has remained largely unrealized. In this work, we tackle the persistent challenges of catastrophic forgetting and temporal interference by introducing a temporal alignment-enhanced SNN framework.

Our approach explicitly preserves spike timing dynamics by integrating a soft-DTW alignment loss with a mechanism for the temporal expansion and contraction of output logits. When coupled with buffer-based experience replay within a ResNet19 backbone, our method effectively closes the performance gap between SNNs and standard ANN-based baselines. Empirical results on MNIST and CIFAR10 confirm that this strategy significantly mitigates forgetting, while ablation studies highlight that precise temporal alignment is central to preserving knowledge. These findings demonstrate that SNNs are valid candidates for scalable class-incremental learning, paving the way for future spike-native pipelines on real-world neuromorphic datasets.
\subsubsection{Future Work.}
We plan to scale STAER to stronger backbones, including transformer-based continual learners and incremental fine-tuning of composable large models~\cite{ilharcoediting,porrellosecond}, building on recent continual ViT techniques such as dynamic token expansion and prompt-based adaptation~\cite{douillard2022dytox,wang2022learning,wang2022dualprompt,smith2023coda}.
We also aim to evaluate on more realistic streams such as Multi-domain Task Incremental Learning (MTIL)~\cite{zheng2023preventing} and MTIL-style continual adaptation of vision-language models~\cite{frascaroli2024clip,yu2024boosting,wang2025lora}, as well as on event-based neuromorphic datasets.

%% file: supp.tex
\clearpage
\section*{Supplementary Material}
\addcontentsline{toc}{section}{Supplementary Material}
\vspace{-1em}
\vspace{2em}

Following the table format  used in the main paper, we show in~\Cref{tab:taskIL} the FAA for TIL protocol. 
\methname consistently improves upon spiking replay baselines across both datasets. In particular, with $T = 4$ we achieve the best SNN performance for all buffer sizes on Sequential-MNIST (up to $99.88\%$) and Sequential-CIFAR10 (up to $96.24\%$), outperforming snn-ER, and snn-DER / snn-DER++, reaching almost the \textit{Joint} upper boundaries. With T = 2, our method remains competitive on Sequential-MNIST and Sequential-CIFAR10, achieving the best SNN performance almost across all buffer sizes.

\include{tables/taskIL_table}

%% file: tables/taskIL_table.tex
\begin{table*}[tb]
\caption{Results on Sequential-MNIST and Sequential-CIFAR10 under TIL.
Bold and underlined values denote the best and second-best results within each SNN block (T=2 and T=4), respectively.}
\label{tab:taskIL}

\centering
\small
\setlength{\tabcolsep}{6pt}
\renewcommand{\arraystretch}{1.2}

\begin{tabular}{lcccccc}
\toprule
\textbf{Method} &
\multicolumn{3}{c}{\textbf{S-MNIST}} &
\multicolumn{3}{c}{\textbf{S-CIFAR-10}} \\
\textit{Buffer size} 
 &
\textit{200} & \textit{500} & \textit{5120} &
\textit{200} & \textit{500} & \textit{5120} \\
\midrule

\rowcolor{green}
\multicolumn{7}{c}{\textbf{ANN without T}} \\

JOINT  & & 99.99 & &  & 98.25 &  \\
SGD    & & 51.71 &  & & 55.15 &  \\
\midrule
ER     & 99.52 & 99.73 & 99.93 & 88.32 & 91.68 & 93.62 \\
DER    & 99.65 & 99.82 & 99.91 & 90.21 & 92.24 & 96.20 \\
DER++  & 99.67 & 99.83 & 99.93 & 91.24 & 93.67 & 97.19 \\

\midrule
\rowcolor{green}
\multicolumn{7}{c}{\textbf{SNN with T=2}} \\

snn-JOINT & & 99.72 & & & 96.05 &  \\
snn-SGD   & & 57.55 & & & 59.37 &  \\
\midrule
snn-ER     & 98.65 & 99.31 & 99.20 & 83.17 & 83.99 & 90.51 \\
snn-DER    & \best{99.28} & 99.26 & 99.55 & 83.49 & 86.57 & 92.70 \\
snn-DER++  & 99.07 & \second{99.62} & \second{99.70} & 84.57 & \best{89.38} & \second{94.24} \\

\midrule
\rowcolor{lightblue}
\textbf{\methname} & \second{99.17} & \best{99.63} & \best{99.71} & \best{84.62} & \second{88.93} & \best{94.28}\\

\midrule
\rowcolor{green}
\multicolumn{7}{c}{\textbf{SNN with T=4}} \\

snn-JOINT & & 99.92 & & & 96.75 &  \\
snn-SGD   & & 63.85 & & & 62.67 &  \\
\midrule
snn-ER     & 98.89 & 99.42 & 99.73 & 83.91 & 84.23 & 90.92 \\
snn-DER    & 99.33 & 99.35 & 99.75 & 85.50 & 88.01 & 93.39 \\
snn-DER++  & \second{99.43} & \second{99.74} & \second{99.77} & \second{86.93} & \second{90.22} & \second{95.03} \\

\rowcolor{lightblue}
\textbf{\methname} & \best{99.45} & \best{99.81} & \best{99.88} & \best{88.12} & \best{90.95} & \best{96.24} \\

\bottomrule
\end{tabular}
\end{table*}